\documentclass[a4paper]{article}

\usepackage{INTERSPEECH2018}

\title{Layer Trajectory LSTM}
\name{Jinyu Li, Changliang Liu, Yifan Gong}
\address{
  Microsoft AI and Research
}
\email{ \{jinyli, chanliu, ygong\}@microsoft.com}

\begin{document}

\maketitle
\begin{abstract}
It is popular to stack LSTM layers to get better modeling power, especially when large amount of training data is available. However, an LSTM-RNN with too many vanilla LSTM layers is very hard to train and there still exists the gradient vanishing issue if the network goes too deep.  This issue can be partially solved by adding skip connections between layers, such as residual LSTM. In this paper, we propose a layer trajectory LSTM (ltLSTM) which builds a layer-LSTM using all the layer outputs from a standard multi-layer time-LSTM. This layer-LSTM scans the outputs from time-LSTMs, and uses the summarized layer trajectory  information for final senone classification. The forward-propagation of time-LSTM and layer-LSTM can be handled in two separate threads in parallel so that the network computation time is the same as the standard time-LSTM. With a layer-LSTM running through layers, a gated path is provided from the output layer to the bottom layer, alleviating  the gradient vanishing issue. Trained  with 30 thousand hours of EN-US Microsoft internal data, the proposed ltLSTM performed  significantly better than the standard multi-layer LSTM and residual LSTM, with up to 9.0\% relative word error rate reduction across different tasks.
\end{abstract}

\section{Introduction}

Recently, significant progress has been made in automatic speech recognition (ASR) when switching from the deep neural networks (DNNs) \cite{DNN4ASR-hinton2012} to recurrent neural networks (RNNs) with long short-term memory (LSTM) units \cite{Hochreiter1997long}, which solve the gradient vanishing or exploding issues in standard RNNs by using input, output and forget gates to achieve a network that can maintain state and propagate gradients in a stable fashion over long spans of time. These LSTM-RNNs  \cite{Graves2013speech, Sak2014long, li2015constructing, Miao15, Miao16} and their variants such as two-dimensional LSTM-RNNs \cite{Li15FLSTM, Li16TFLSTM, sainath2016modeling} have been shown to outperform DNNs on a variety of ASR tasks. 

It is popular to stack multiple LSTM layers to get better modeling power \cite{Sak2014long}, especially when large amount of training data is available. However, an LSTM-RNN with too many vanilla LSTM layers is very hard to train and there still exists the gradient vanishing issue if the network goes too deep \cite{HighwayBLSTM-zhang2016, hsu2016prioritized}.  This issue can be partially solved by adding skip connections or gating functions between layers. 

Residual LSTM \cite{zhao2016multidimensional, kim2017residual} uses shortcut connections between LSTM layers, and hence  provides a way to alleviate the gradient vanishing problem. In the highway LSTM \cite{HighwayBLSTM-zhang2016}, memory cells of adjacent layers are connected by gated direct links which provide a path for information to flow between layers more directly without decay. Therefore, it alleviates the gradient vanishing issue and enables the training of much deeper LSTM-RNN networks.  In \cite{pundak2017highway},  highway LSTM was investigated with large scale of training data, but only very limited gain was obtained over the standard multi-layer LSTM. Grid LSTM \cite{kalchbrenner2015grid} is a more general LSTM which arranges the LSTM memory cells into a multidimensional grid along both time and layer axis. It was extended in \cite{hsu2016prioritized, najafian2017automatic} as prioritized grid LSTM which was shown to outperform highway LSTM on several ASR tasks. 

All the aforementioned models work in a layer-by-layer and step-by-step fashion. The output of a LSTM unit (either the standard time LSTM or grid LSTM) is used as the input of the LSTM at the same time step in the next layer and the recurrent input of the LSTM at the next time step in the same layer. The output of the highest layer LSTM is used for final senone (tied triphone states) classification. However, it may not be optimal that the LSTM outputs serve the purpose of both recurrence along time axis (for temporal modeling) and senone classification along the layer axis (for target discrimination). 

In this paper, we decouple the purposes of time recurrence and senone classification by proposing a layer trajectory LSTM (ltLSTM) which builds a layer-LSTM using the outputs from all the layers of a standard multi-layer time-LSTM.  The time-LSTM is used for temporal modeling with time recurrence, while the layer-LSTM scans the outputs from multiple time-LSTM layers and uses the summarized layer trajectory information for final senone classification. Hence, the forward-propagation of the time-LSTM in next frame is independent of the calculation of the layer-LSTM in current frame, therefore the evaluation of time-LSTM and layer-LSTM can be handled in two separate threads in parallel and the network computational time can be kept the same as the standard time-LSTM. With a layer-LSTM running through layers, a gated path is provided from the output layer to the bottom layer, reducing the gradient vanishing issue. We evaluate the proposed method by training various models with  30 thousand (k) hours of EN-US data which pools from  Microsoft Cortana, Conversation, and xBox data with mixed close-talk and far-field utterances. The proposed ltLSTM is significantly better than the standard multi-layer LSTM and residual LSTM. 

The rest of the paper is organized as follows. In Section \ref{sec:ltLSTM}, we explore different LSTM structures: standard multi-layer LSTM, Residual LSTM, and the proposed ltLSTM. We also propose a new way to reduce the computational cost of LSTM by factorizing the gates calculation. We evaluate the proposed models in Section \ref{sec:exp}, and conclude our study in Section \ref{sec:con}.

%\section{Layer trajectory LSTM}
\section{Exploring LSTM structures}
\label{sec:ltLSTM}

In this section, we first introduce the standard multi-layer LSTM and residual LSTM (ResLSTM). Then, we describe our proposed layer trajectory LSTM. Finally, a factorized gate LSTM is proposed to reduce the computational cost of LSTM units. 

\subsection{LSTM}
\label{ssec:LSTM}

\begin{figure}[t]
  \centering
  \includegraphics[width=\linewidth]{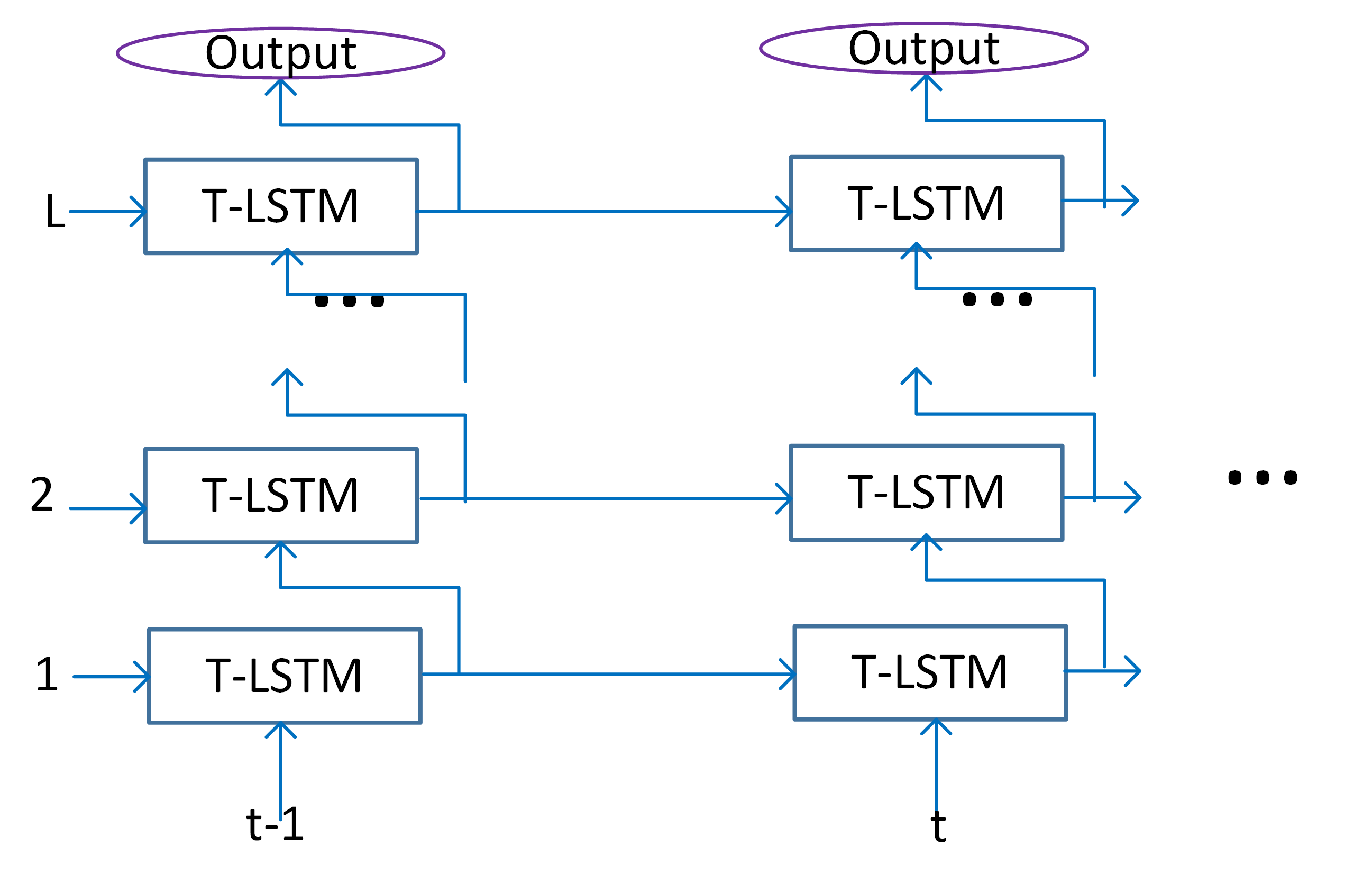}
  \caption{Flowchart of multi-layer time-LSTM (T-LSTM). The output of a T-LSTM is used as the input of the T-LSTM at the same time step in the next layer and the recurrent input of the T-LSTM at the next time step in the same layer.}
  \label{fig:tLSTM}
\end{figure}

The standard LSTM is a time-LSTM which does temporal modeling via time recurrence by taking the output of time-LSTM at previous time step as the input of the time-LSTM at current time step. To increase modeling power, multiple layers of LSTM units are stacked together to form a multi-layer LSTM which is shown in Figure~\ref{fig:tLSTM}. At  time step $t$, the vector formulas of the computation of the $l$-th layer LSTM units can be described as:
  \begin{align}
  \textbf{i}_t^l &= \sigma ( \textbf{W}^l_{ix} \textbf{x}_{t}^l + \textbf{W}^l_{ih} \textbf{h}_{t-1}^l + \textbf{p}^l_{i} \odot \textbf{c}_{t-1}^l + \textbf{b}^l_{i}) \label{eq:igate}\\
  \textbf{f}_t^l &= \sigma ( \textbf{W}^l_{fx} \textbf{x}_{t}^l + \textbf{W}^l_{fh} \textbf{h}_{t-1}^l + \textbf{p}^l_{f} \odot \textbf{c}_{t-1}^l + \textbf{b}^l_{f}) \label{eq:fgate}\\
  \textbf{c}_t^l &= \textbf{f}_t^l \odot \textbf{c}_{t-1}^l + \textbf{i}_t^l \odot \phi( \textbf{W}^l_{cx} \textbf{x}_{t}^l + \textbf{W}^l_{ch} \textbf{h}_{t-1}^l + \textbf{b}^l_{c}) \label{eq:cell} \\
  \textbf{o}_t^l &= \sigma ( \textbf{W}^l_{ox} \textbf{x}_{t}^l + \textbf{W}^l_{oh} \textbf{h}_{t-1}^l + \textbf{p}^l_{o} \odot \textbf{c}_{t}^l + \textbf{b}^l_{o}) \label{eq:ogate} \\
  \textbf{h}_t^l &= \textbf{o}_{t}^l \odot \phi(\textbf{c}_t^l) \label{eq:out}
  \end{align}
where $\textbf{x}_{t}^l$ is the input vector for the $l$-th layer with
 \begin{equation}
    \textbf{x}_{t}^l = 
\begin{cases}
    \textbf{h}_{t}^{l-1},& \text{if } l > 1 \label{eq:lstmx} \\
    \textbf{s}_t,              & \text{if } l = 1
\end{cases}
\end{equation}
$\textbf{s}_t$ is the speech spectrum input at time step $t$. The vectors  $\textbf{i}_t^l$, $\textbf{o}_t^l$, $\textbf{f}_t^l$, $\textbf{c}_t^l$ are the activation of the input, output, forget gates, and memory cells, respectively. $\textbf{h}_t^l$ is the output of the time-LSTM.  $\textbf{W}^l_{.x}$ and  $\textbf{W}^l_{.h}$ are the weight matrices for the inputs $\textbf{x}_{t}^l$ and the recurrent inputs $\textbf{h}_{t-1}^l$, respectively. $\textbf{b}^l_{.}$ are bias vectors. $\textbf{p}^l_{i}$, $\textbf{p}^l_{o}$, $\textbf{p}^l_{f}$ are parameter vectors associated with peephole connections. The functions $\sigma$ and $\phi$ are the logistic sigmoid and hyperbolic tangent nonlinearity, respectively. The operation $\odot$ represents element-wise multiplication of vectors.

From Figure~\ref{fig:tLSTM}, we can see that the output of a time-LSTM is used as the input of the time-LSTM at the same time step in the next layer and the recurrent input of the time-LSTM at the next time step in the same layer. The last hidden layer's output is used to predict senone labels for senone classification. Therefore, the same output is used for the purpose of temporal model along time axis and the purpose of target discrimination along the layer axis. However, these two purposes are indeed very different. Hence, the standard time-LSTM modeling may not be optimal. 

\subsection{Residual LSTM}
\label{ssec:ResLSTM}
Similar to Residual CNN \cite{RESNET-he2015} which recently achieves great success in the image classification task, residual LSTM (ResLSTM) is very straightforward with the direct shortcut path across layers by changing Eq. \eqref{eq:lstmx} to Eq. \eqref{eq:reslstmx} so that gradient vanishing issue can be partially solved. 
 \begin{equation}
    \textbf{x}_{t}^l = 
\begin{cases}
    \textbf{x}_{t}^{l-1} +  \textbf{h}_{t}^{l-1},& \text{if } l > 1 \label{eq:reslstmx} \\
    \textbf{s}_t,              & \text{if } l = 1
\end{cases}
\end{equation}
We will use ResLSTM as a baseline model with skip connection in Section \ref{sec:exp}.

Although ResLSTM can partially solve the gradient vanishing issue, it still has the same challenges as the standard time-LSTM -- the output vector works for two very different purposes: temporal modeling and senone classification. 

\subsection{Layer trajectory LSTM}
\label{ssec:ltLSTM}

\begin{figure}[t]
  \centering
  \includegraphics[width=\linewidth]{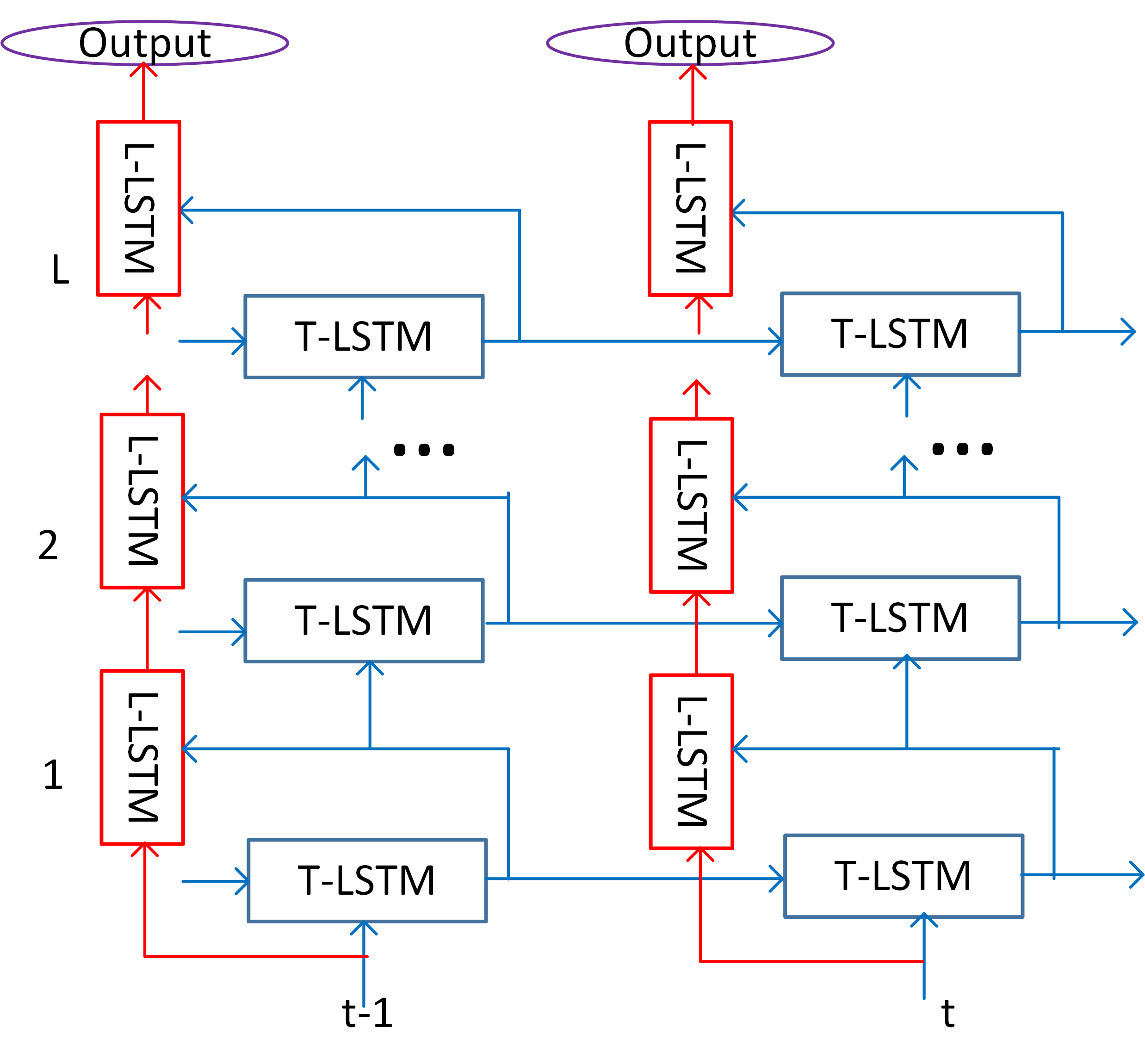}
  \caption{Flowchart of layer trajectory LSTM (ltLSTM). 
	%The output of a T-LSTM is still used as the input of the T-LSTM at the current time step in the next layer and the recurrent input of the T-LSTM at the next time step in the same layer. 
	Layer-LSTM (L-LSTM) is used to scan the outputs of time-LSTM (T-LSTM) across all layers at the current time step to get summarized layer trajectory information for senone classification. 
	%The recurrence of L-LSTM is across layers. 
	There is no time recurrence for L-LSTM. Time recurrence only exists between T-LSTMs at different time steps.}
  \label{fig:ltLSTM}
\end{figure}

As discussed above, it may not be optimal that the output of time-LSTM serves both the purposes of temporal modeling and senone classification. In this study, we decouple these two purposes by proposing a layer trajectory LSTM (ltLSTM) which builds a layer-LSTM using the outputs from all the time-LSTM layers, shown in  Figure ~\ref{fig:ltLSTM}. The weights are not shared between layers because sharing doesn't bring any computational benefit. 
The time-LSTM is used for the purpose of temporal modeling via time recurrence, while the layer-LSTM scans the outputs from multiple time-LSTM layers and uses the summarized layer trajectory information for final senone classification.
% because we believe the multi-layer outputs from time-LSTM carry more information for classification than the output from the last time-LSTM layer only. 
With a layer-LSTM running through layers, a gated path is provided from the output layer to the bottom layer, reducing the gradient vanishing issue.

In ltLSTM, the formulation of time-LSTM is still the standard LSTM formulation, with Eqs. \eqref{eq:igate} -- \eqref{eq:out}. As shown in Figure ~\ref{fig:ltLSTM}, there is no time recurrence between layer-LSTMs across different time steps. Hence, the formulation of layer-LSTM only has the recurrence across layers as:
 \begin{align}
 \textbf{j}_t^l &= \sigma ( \textbf{U}^l_{jh} \textbf{h}_{t}^l + \textbf{U}^l_{jg} \textbf{g}_{t}^{l-1} + \textbf{q}^l_{j} \odot \textbf{m}_{t}^{l-1} + \textbf{d}^l_{j}) \label{eq:layer_igate}\\
 \textbf{e}_t^l &= \sigma ( \textbf{U}^l_{eh} \textbf{h}_{t}^l + \textbf{U}^l_{eg} \textbf{g}_{t}^{l-1} + \textbf{q}^l_{e} \odot \textbf{m}_{t}^{l-1} + \textbf{d}^l_{e}) \label{eq:layer_fgate}\\
 \textbf{m}_t^l &= \textbf{e}_t^l \odot \textbf{m}_{t}^{l-1} + \textbf{j}_t^l \odot \phi( \textbf{U}^l_{sh} \textbf{h}_{t}^l + \textbf{U}^l_{sg} \textbf{g}_{t}^{l-1} + \textbf{d}^l_{s}) \label{eq:layer_cell} \\
 \textbf{v}_t^l &= \sigma ( \textbf{U}^l_{vh} \textbf{h}_{t}^l + \textbf{U}^l_{vg} \textbf{g}_{t}^{l-1} + \textbf{q}^l_{v} \odot \textbf{m}_{t}^l + \textbf{d}^l_{v}) \label{eq:layer_ogate} \\
 \textbf{g}_t^l &= \textbf{v}_{t}^l \odot \phi(\textbf{m}_t^l) \label{eq:layer_out}
 \end{align}
The vectors  $\textbf{j}_t^l$, $\textbf{v}_t^l$, $\textbf{e}_t^l$, $\textbf{m}_t^l$ are the activation of the input, output, forget gates, and memory cell of the layer-LSTM, respectively. $\textbf{g}_t^l$ is the output of the layer-LSTM.  The matrices $\textbf{U}^l_{.h}$ and  $\textbf{U}^l_{.g}$ terms are the weight matrices for the inputs $\textbf{h}_{t}^l$ and the recurrent inputs $\textbf{g}_{t}^{l-1}$, respectively. The $\textbf{d}^l_{.}$ are bias vectors.  The $\textbf{q}^l_{j}$, $\textbf{q}^l_{v}$, $\textbf{q}^l_{e}$ are parameter vectors associated with peephole connections.

Comparing Eqs. \eqref{eq:igate} -- \eqref{eq:out} with Eqs. \eqref{eq:layer_igate} -- \eqref{eq:layer_out}, we can see the biggest difference is the recurrence now happens across the layers with $\textbf{g}_{t}^{l-1}$ in layer-LSTM, compared to the time recurrence with $\textbf{h}_{t-1}^l$ in time-LSTM. Layer-LSTM uses the output of time-LSTM at current layer, $\textbf{h}_{t}^l$, as the input, compared to the $\textbf{x}_{t}^l$ in time-LSTM. 

It is a common practice to deploy a complicated model by reducing parallel computational time, e.g., \cite{li2017reducing}. Because the forward-propagation of  the time-LSTM at next time step is independent of the calculation of the layer-LSTM at current time step, the forward-propagation of time-LSTM and layer-LSTM can be handled in two separate threads in parallel and the network computational time is the same as the standard time-LSTM which operates in a layer-by-layer and frame-by-frame fashion.  Another advantage of decoupling the time and layer operation in ltLSTM is that layer-LSTM can be evaluated with batching \cite{vanhoucke2011improving} which was proposed to improve the runtime of feed-forward DNNs by evaluating the network scores from several time frames at the same time. However, batching cannot be applied to standard time-LSTM because the input of current frame is from the output of previous frame. Since there is no time recurrence between layer-LSTM across different frames, batching can be applied to evaluate layer-LSTM once the time-LSTM vectors have been calculated in multiple frames. 

\subsection{Comparison with grid LSTMs}
\label{ssec:gLSTM}
Grid LSTM (gLSTM) \cite{kalchbrenner2015grid} and prioritized grid LSTM (pgLSTM) \cite{hsu2016prioritized, najafian2017automatic} can be considered as a multidimensional LSTM which arranges the LSTM memory cells along both time and layer axis.  They modify the LSTM units with multi-dimensional formulation and still process the speech input in a step-by-step and layer-by-layer fashion. The operation along time and layer dimensions is mixed together. In contrast, ltLSTM decouples the jobs of temporal modeling with time-LSTM and target classification with layer-LSTM. The evaluation of time-LSTM doesn't rely on the value of layer-LSTM in previous time step or lower layer. Hence, ltLSTM enjoys clear computational advantage as discussed in the previous section. Because of the function decoupling, it is not necessary to use LSTM units for modeling layer dependency. We can just replace layer-LSTM units with layer-DNN units or any other units, which gives more modeling flexibility \cite{Li18gltLSTM}.

\subsection{Factorized gate  LSTM}
\label{ssec:FG-LSTM}

The computational cost of LSTM is always a concern. There are lots of attempts  \cite{Yu-RecentProgDeepLearningAcousticModels} to reduce the computational cost, such as getting low-rank matrices with  singular value decomposition (SVD) \cite{Xue13, prabhavalkar2016compression}, model compression via teacher-student (T/S) learning \cite{li2014learning} or knowledge distillation \cite{hinton2015distilling}, scalar quantization  \cite{vanhoucke2011improving}, and vector quantization \cite{wang2015small} etc. The computational cost can also be reduced by exploring different model structures \cite{Miao16, greff2017lstm} or using lower frame rate strategies \cite{Miao16, pundak2016lower}. 

In this section, we focus on reducing the size of weight matrices used to calculate input, output, and forget gates in the LSTM unit. Those matrices usually are of large size, resulting the major computational cost in the network evaluation. For example, some typical LSTM-RNN systems usually have around 1024 memory cells \cite{Sak2014long, Li2018Speaker} in the LSTM unit, which means the dimension of gate vectors is 1024. Usually a linear projection layer is applied to the LSTM output vector to reduce its dimension, for example to 512. Hence the two weight matrices , $\textbf{W}^l_{ix}$ and $\textbf{W}^l_{ih}$, used to calculate the input gate vector in Eq. \eqref{eq:igate} are of dimension $1024 \text{x} 512$. 
 
In Eq. \eqref{eq:fg-input}, we factorize the input gate vector calculation as the square root of the product of two vectors, ${\acute{\bf{i}}}_t^l$ and $\{{\grave{\bf{i}}}_t^l\}^T$, which are calculated by Eqs. \eqref{eq:i1} and \eqref{eq:i2}. $vec(.)$ is the operation that squashes a $k\text{x}k$ matrix into a $m$-dimension vector, where $m=k\text{x}k$. Peephole connections are not used in  Eqs. \eqref{eq:i1} and \eqref{eq:i2} because of dimension mismatch between the state vector ($m$ dimension) and factorized gate vector ($k$ dimension). 
\begin{align}
	{\acute{\bf{i}}}_t^l &= \sigma ( \acute{\bf{W}}^l_{ix} \textbf{x}_{t}^l + \acute{\bf{W}}^l_{ih} \textbf{h}_{t-1}^l + \acute{\bf{b}}^l_{i}) \label{eq:i1}\\
	{\grave{\bf{i}}}_t^l &= \sigma ( \grave{\bf{W}}^l_{ix} \textbf{x}_{t}^l + \grave{\bf{W}}^l_{ih} \textbf{h}_{t-1}^l  + \grave{\bf{b}}^l_{i}) \label{eq:i2} \\
	\hat{\bf{i}}_t^l &= vec ( sqrt ( {\acute{\bf{i}}}_t^l ~~ \{{\grave{\bf{i}}}_t^l\}^T)	) \label{eq:fg-input}
	%\hat{\bf{i}}_t^l &= sqrt \{ vec\{ {\acute{\bf{i}}}_t^l ~~ \{{\grave{\bf{i}}}_t^l\}^T\}	\} \label{eq:fg-input}
\end{align}

For the above example, instead of having two $1024 \text{x} 512$ matrices for the input gate calculation in Eq. \eqref{eq:fg-input}, only four  $32 \text{x} 512$ matrices ($1024 = 32 \text{x} 32$) are involved in Eqs. \eqref{eq:i1} and \eqref{eq:i2}. The computation cost is reduced to only 1/16 for the input gate vector calculation. 

Similar formulations can be applied to the forget and output gates of the time-LSTM in Eqs. \eqref{eq:fgate} and \eqref{eq:ogate}, and the input, forget, and output gates of layer-LSTM in Eqs. \eqref{eq:layer_igate}, \eqref{eq:layer_fgate} and \eqref{eq:layer_ogate}. %The proposed factorized gate LSTM can be jointly used with aforementioned runtime cost reduction technologies such as SVD, T/S learning, or quantization. 

\section{Experiments}
\label{sec:exp}

We compare standard multi-layer LSTM, ResLSTM, and the proposed ltLSTM in this section. All these models use standard LSTM units as basic building blocks, different from grid LSTM which modifies the LSTM units with multi-dimensional formulation. All models were trained with 30 thousand (k) hours of anonymized and  transcribed Microsoft production data, including Cortana \cite{Li2018Speaker}, xBox \cite{liu2015svd}, and  Conversation data, which is a mixture of close-talk and far-field utterances from a variety of devices. The first model was built as a 4-layer LSTM-RNN with projection layer as what we usually did \cite{Li2018Speaker}. Each LSTM layer has 1024 hidden units and the output size of each LSTM layer is reduced to 512 using a linear projection layer. The output layer has 9404 nodes, modeling the senone labels. The target senone label is delayed by 5 frames as in \cite{Sak2014long}. The input feature is 80-dimension log Mel filter bank.  We applied frame skipping \cite{Miao16} to reduce the runtime cost. Note that in this study, we only compare the baseline full-rank cross-entropy models. If we want to deploy models into production, we will further apply SVD training \cite{Xue13} and sequence discriminative training using the maximum mutual information (MMI) criterion with F-smoothing \cite{su2013error}, as the systems described in \cite{Li2018Speaker}. The LM is a 5-gram with around 100 million (M) ngrams.  

We evaluate all models with  Cortana and Conversation test sets. Both sets contain mixed close-talk and far-field utterances, with 439k and 111k words, respectively. The Cortana test set has shorter utterances related to voice search and commands, while the Conversation test set has longer utterances for conversation. As shown in Table \ref{tab:wer}, the 4-layer LSTM model obtained 10.37\% and 19.41\% WER on these 2 test sets, respectively. 

\begin{table}[t]
  \caption{WERs of LSTM, ResLSTM, and ltLSTM models on Cortana and Conversation test sets. Both test sets are mixed with close-talk and far-field utterances.}
  \label{tab:wer}
  \centering
  \begin{tabular}{l|c|c}
    \toprule
							& 	Cortana				 & 			Conversation              \\							
    \midrule
		4-layer LSTM 			& 10.37 & 19.41 \\
    6-layer LSTM      & 9.85 & 19.20         \\
		10-layer LSTM     &	10.58 & 19.92 \\ \hline
		
		6-layer ResLSTM	 & 9.99 & 18.85 \\
		10-layer ResLSTM & 9.68 & 18.15 \\	\hline
		6-layer ltLSTM   & \textbf{9.28} & \textbf{17.47} 	 \\

    \bottomrule
  \end{tabular}
\end{table}

Then, we simply increased the number of LSTM layers to 6 and 10. Increasing from 4 layers to 6 layers, the multi-layer LSTM got improvement across all tasks, with 9.85\% and 19.20\% WERs on Cortana and Conversation test sets, respectively. However, when increasing to 10 layers, the multi-layer LSTM got lots of degradation, consistent with the observations in literature \cite{HighwayBLSTM-zhang2016, hsu2016prioritized}. 
% We will use the 6-layer LSTM as the baseline for all the results comparison.

The 6-layer ResLSTM obtained very similar WERs as the 6-layer LSTM, with improvement on Conversation test sets, but slight degradation on Cortana test sets. However, different from the behavior of the 10-layer LSTM, consistent improvement was obtained with the 10-layer ResLSTM which got 9.68\% and 18.15\% WERs on Cortana and Conversation test sets, respectively. This clearly demonstrates the effectiveness of skipping connection for reducing the gradient vanishing issue.

Finally, the 6-layer ltLSTM got significant improvement over all  models, obtaining 9.28\% and 17.47\% WERs on Cortana and Conversation test sets, respectively. This represents 5.8\% and 9.0\% relative WER reduction from the 6-layer LSTM, or 4.1\% and 3.7\% relative WER reduction from the 10-layer ResLSTM on Cortana and Conversation test sets, respectively.

\begin{table}[t]
  \caption{Total and parallel per-thread computational costs of LSTM, ResLSTM, and ltLSTM models in terms of million (M) operations per frame.}
  \label{tab:cost}
  \centering
  \begin{tabular}{l|c|c}
    \toprule
    %\textbf{Model}      & \textbf{Entities in a Paper}                \\
							&  Total (M) & Parallel per thread (M)\\
    \midrule
		4-layer LSTM 			& 22 & 22  \\
    6-layer LSTM      & 31 & 31     \\
		10-layer LSTM     & 49 & 49	\\ \hline
		
		6-layer ResLSTM	 & 31 & 31  \\
		10-layer ResLSTM & 49 & 49 \\	\hline
		6-layer ltLSTM   & \textbf{57} & \textbf{31}  \\
    \bottomrule
  \end{tabular}
\end{table}

In Table \ref{tab:cost}, we examine the total and parallel computational costs of all LSTM, ResLSTM, and ltLSTM models.  Both LSTM and ResLSTM operate in a frame-by-frame and layer-by-layer fashion, therefore the total and parallel computational costs are same. As described in Section \ref{ssec:ltLSTM}, the layer-LSTM and time-LSTM inside ltLSTM can be evaluated in parallel as there is no time recurrence between layer-LSTMs at different time steps. As a result, the parallel computational cost is about 31 M per frame, which is the same as that of the 6-layer LSTM. 

\begin{table}[t]
  \caption{WERs of the 6-layer ltLSTM and its factorized gate versions on Cortana and Conversation test sets. Both test sets are mixed with close-talk and far-field utterances. }
  \label{tab:wer-fac}
  \centering
  \begin{tabular}{l|c|c}
    \toprule
						& 	Cortana				 & 			Conversation              \\		                          
							   \midrule
		 full  &  \textbf{9.28} & \textbf{17.47} 	 \\
		factorized input   & 9.62 & 18.31	 \\
		factorized output   & 9.50 & 18.11	 \\	%done
		factorized forget   & 9.57 & 17.97	 \\	%done
		%factor forget couple (done)  & 11.89 & 6.90 & 20.73 & 16.01 \\
    \bottomrule
  \end{tabular}
\end{table}

We applied factorized gate LSTM described in Section \ref{ssec:FG-LSTM} to the 6-layer ltLSTM and evaluated the method in Table \ref{tab:wer-fac}. We factorized input gates in both time and layer LSTMs by reducing the calculation of a 1024-dimension gate vector into the calculation of two 32-dimension gate vectors as in Eqs. \eqref{eq:i1}, \eqref{eq:i2},   and \eqref{eq:fg-input}. We also applied similar operation to factorize output and forget gates in  both time and layer LSTMs.  All the factorized gate operation increased WER. Clearly, no magic happens even with the factorization in Eq.\eqref{eq:fg-input} because two 32-dimension gate vectors carry much less information than what a 1024-dimension gate vector can carry. The impact of factorizing forget gate is the smallest, with relative 2.3\% and 3.7\% WER increase from the full version of ltLSTM without any factorization, although it is still better than all the LSTM and ResLSTM models. Factorizing input gates has the biggest degradation. Given the loss, we didn't evaluate the setup which factorizes all the gates together.
With single gate factorization, the parallel computational cost is reduced to 25 M operation per frame which is even lower than that of the 6-layer LSTM or ResLSTM while the WER of factorized gate ltLSTM is clearly better than that of the 6-layer LSTM or ResLSTM. 
%\begin{table}[t]
  %\caption{Total and parallel  per thread computational costs of the 6-layer ltLSTM and its factorized gate versions in terms of million (M) operations per frame.}
  %\label{tab:cost-fac}
  %\centering
  %\begin{tabular}{l|c|c}
    %\toprule
    %%\textbf{Model}      & \textbf{Entities in a Paper}                \\
							%&  Total (M) & Parallel  per thread (M)\\
    %\midrule
		%full  & 113 & 61  \\
		%factorized input & 92 & 50 \\
		%factorized output & 92 & 50 \\
		%factorized forget & 92 & 50 \\
    %\bottomrule
  %\end{tabular}
%\end{table}

% In Table \ref{tab:cost-fac}, we list the total and parallel computational costs of the 6-layer ltLSTM and its factorized gate versions. Even with single gate factorization, the parallel computational cost is reduced to 50 M operation per frame which is even lower than that of the 6-layer LSTM or ResLSTM while the WER of factorized gate ltLSTM is clearly better that of the 6-layer LSTM or ResLSTM. 

\section{Conclusions and Future Works}
\label{sec:con}

In this paper, we proposed a novel model called ltLSTM which scans the outputs of the multi-layer time-LSTM with a layer-LSTM to learn layer trajectory information which is used for classification. This model decouples the tasks of temporal modeling and target classification by using time-LSTM and layer-LSTM, respectively. It brings the benefits of both accuracy and runtime. Trained with 30k hours of speech data, the 6-layer ltLSTM  improves the baseline 6-layer LSTM with relative 5.8\% and 9.0\% WER reduction on Cortana and Conversation test sets respectively and reduces the WERs of the 10-layer ResLSTM by 4.1\% and 3.7\% relative. With parallel computation, the model evaluation time of the 6-layer ltLSTM is kept the same as that of the 6-layer LSTM. Furthermore, we proposed to factorize gates inside LSTM units to reduce the runtime cost. Applied to the 6-layer ltLSTM, the model has smaller parallel computational cost and better accuracy than that of the 6-layer LSTM or ResLSTM. 

Recently, we blended attention mechanism \cite{Attention-bahdanau2014} into CTC  modeling and achieved very good accuracy improvement \cite{Das18CTCAttention, Li18CTCnoOOV}. We are now using similar idea to further improve ltLSTM. As noted in Section \ref{ssec:gLSTM}, it is not necessary to use LSTM units for modeling layer dependency. We are working on a generalized ltLSTM which can employ any units for  modeling layer dependency. All these works will be reported in \cite{Li18gltLSTM}. 

\bibliographystyle{IEEEtran}

\bibliography{mybib}

\end{document}